\documentclass[conf]{new-aiaa}
\usepackage[utf8]{inputenc}

\usepackage{tikz}
\usepackage{pgfplots}
\usepgfplotslibrary{colorbrewer}
\usepackage{standalone}
\usepackage{graphicx}
\usepackage{amsmath}
\usepackage{cleveref}
\usepackage{booktabs}
\usepackage{rotating}
\usetikzlibrary{arrows.meta}
\usetikzlibrary{arrows}
\tikzset{>=stealth'}
\usepackage{siunitx}
\usepackage{longtable,tabularx}
\title{Visual Depth Mapping from Monocular Images \\ using Recurrent Convolutional Neural Networks}
\newcommand\samefootnote[1][\value{footnote}]{\footnotemark[#1]}
\author{John Mern,\footnote{Graduate Student Researcher, Stanford Intelligent Systems Laboratory, and AIAA Student Member} Kyle Julian,\samefootnote\ Rachael E. Tompa,\samefootnote\ and Mykel J. Kochenderfer\footnote{Assistant Professor, Stanford Intelligent Systems Laboratory, and AIAA Associate Fellow}}
\affil{Department of Aeronautics and Astronautics, Stanford University, Stanford, CA, 94305}

\begin{document}

\maketitle
\begin{abstract}
A reliable sense-and-avoid system is critical to enabling safe autonomous operation of unmanned aircraft. 
Existing sense-and-avoid methods often require specialized sensors that are too large or power intensive for use on small unmanned vehicles. 
This paper presents a method to estimate object distances based on visual image sequences, allowing for the use of low-cost, on-board monocular cameras as simple collision avoidance sensors. 
We present a deep recurrent convolutional neural network and training method to generate depth maps from video sequences. 
Our network is trained using simulated camera and depth data generated with Microsoft's AirSim simulator.
Empirically, we show that our model achieves superior performance compared to models generated using prior methods. 
We further demonstrate that the method can be used for sense-and-avoid of obstacles in simulation. 
\end{abstract}

\section{Introduction}

Effective sense-and-avoid systems are necessary to safely integrate unmanned aircraft into the airspace~\cite{valavanis2015}.
Many systems require specialized sensors and extensive computational resources creating the challenge of adhering to aircraft size, weight, and power (SWaP) constraints~\cite{Sahawneh2016}.
Embedded digital cameras, such as those commonly installed in cell phones, are common low-SWaP sensors that can be easily accommodated on-board most small unmanned aircraft.

Camera images cannot be used directly for sense-and-avoid because they do not provide the three-dimensional location of potential obstacles.  
In this paper, we present a method to estimate three-dimensional locations using image sequences from simple monocular cameras. 
Our method generates a relative depth map of each pixel in a camera field-of-view (FoV) in the direction normal to the image plane. 
The resulting depth maps can then be used in a variety of applications, such as Simultaneous Localization and Mapping (SLAM) or sense-and-avoid. 
This method does not require specialized sensors allowing it to be used on SWaP-constrained vehicles where other systems are infeasible. 

The proposed method uses a deep neural network to map visual image sequences to corresponding relative depth maps.
In order to account for the correlation between sequential input frames, we propose a recurrent convolutional neural network (R-CNN) architecture~\cite{Pinheiro2013}.
We present this general architecture and recommend an auto-encoder design based on convolutional Gated Recurrent Units (C-GRUs). In addition, we present a method to effectively train the network over image sequences using stochastic mini-batches.

We demonstrate the effectiveness of the depth extraction approach in Microsoft's AirSim simulator~\cite{Shah2017}. Using AirSim, we generate matched-pair sets of images from a simulated on-board camera and the depth map of the scene in the field of view.   
We provide qualitative examples of the depth maps generated by our method and quantitative evaluations using conventional metrics from the field of computer vision. 
Our method outperforms three previously proposed deep learning-based methods. 
We also show that the accuracy of the depth maps generated by our approach is sufficient for sense-and-avoid of stationary obstacles without additional sensor or telemetry data. 

\section{Prior Work}
Prior attempts to extract depth maps from images have employed a variety of approaches. 
Early approaches were based on geometric inferencing using stereo images~\cite{Lucas1981}. 
Such methods used a pair of cameras fixed at a known displacement from one-another that collect a set of stereo images. 
Distance could then be estimated by extracting features from objects in the scene and calculating the depth from the disparity of the location of the feature in each camera view. 
These methods are limited in how well they can resolve depth at different distances~\cite{Alvertos1988}.

Recent developments have improved the performance of stereoscopic techniques~\cite{Heo2013}. 
These methods continue to be limited in the resolution they can accurately provide in many environments where feature-extraction is challenging and are often sensitive to camera performance~\cite{Im2015}.
Application of deep neural networks to stereoscopic pairs have been shown to alleviate some of these challenges~\cite{Luo2016,Zbontar2016}. 

Conventional computer vision methods have been proposed to extract depth from single images using geometric features. 
These methods generally rely on heuristic techniques that require the presence of strong scene features such as vanishing perspective lines and clear horizon~\cite{Battiato2004,Battiato2004a,Tsai2006}. 
Such methods have achieved reasonable success in indoor environments~\cite{Silberman2012} where such features are commonly present, however, this success has not generalized to outdoor environments. 
Some researchers have used multi-scale Markov random fields to model the relationship between features and depth~\cite{saxena2008}. 

Recent techniques have used deep convolutional neural networks (CNNs) to build depth maps from single images~\cite{Laina2016,Eigen2014,liu2015}. 
These techniques often require extensive pre- or post-processing~\cite{Garg2016}. 
While the performance of the different methods vary, these methods tend to produce accurate resolution of low-frequency information. 

Additional methods have been introduced to extract relative scene information from video sequences. 
Geometric methods based on feature-displacement and optical flow have been used on videos for depth mapping~\cite{Chang2007}. 
Like the geometric methods used on single images, these methods are dependent on scene features in order to provide accurate estimates and are therefore limited in the environments for which they can provide accurate resolution. 

Our work draws several insights  from within the field of deep computer vision. 
In particular, we draw from the work done with deep auto-encoders. 
Deep auto-encoders use CNNs to generate compressed representations of an input image and the mapping of that compressed representation back to the original image~\cite{vincent2010}. 
Unlike CNNs, which can be used to map input images to symbolic outputs such as classification labels, auto-encoders train on mappings from images to images. 
The design and training methods for these networks have been successfully applied to various image-to-image translation tasks, such as those in the field of medical imaging~\cite{Ronnenberger2015}. 
Our work casts the depth-mapping problem as an image-to-image translation problem and builds on on these methods.  

\section{Technical Approach}
The main contribution of this work is the introduction of a recurrent convolutional neural network (CNN) for the generation of depth maps from visual image sequences. 
CNNs are a special class of neural network, which are commonly used in machine learning because of their ability to approximate a wide variety of functions~\cite{kuurkova1992}. 
The network is composed of layers with weights and biases that define a mapping from an input to output. 
The outputs of each layer pass through a non-linear activation function before being used as the input for the following layer. 
The weights and biases are optimized to minimize the empirical loss over a training data set.

\begin{figure}
\centering
\includegraphics[width=0.85 \textwidth]{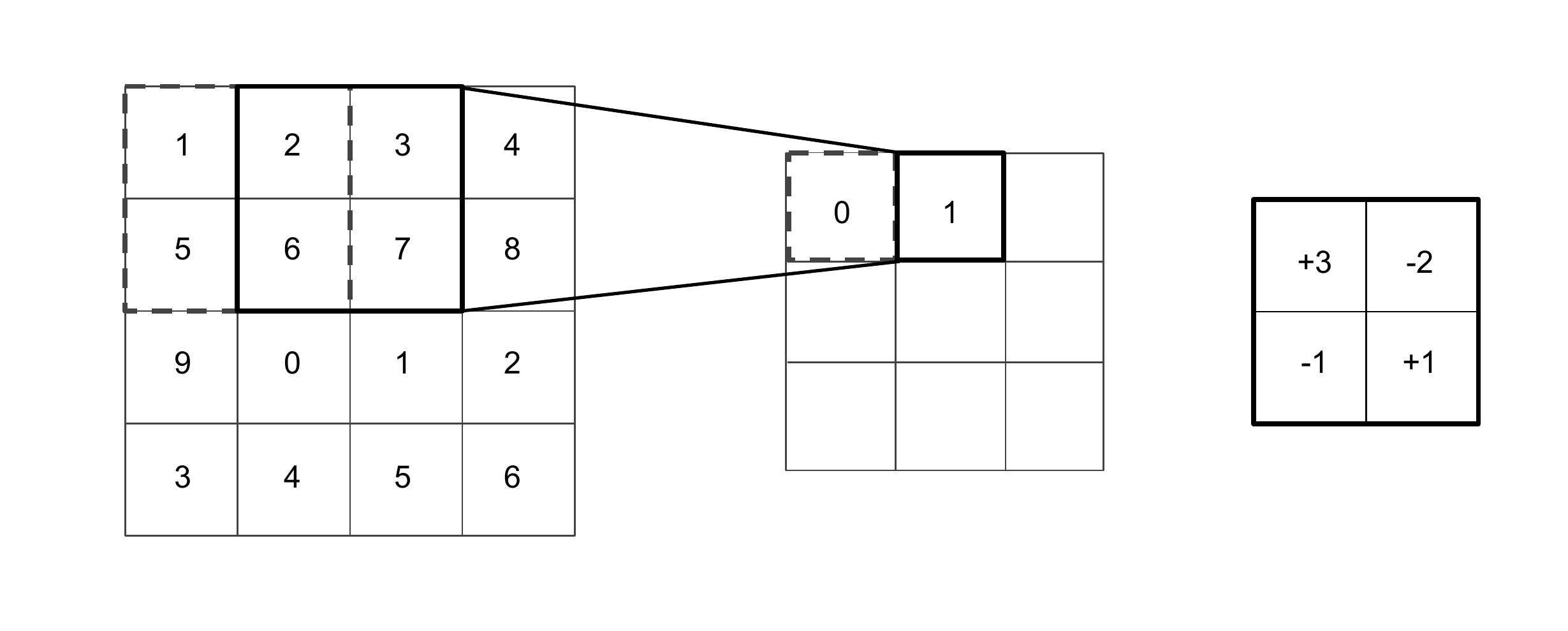}
\caption{Convolutional Filter Example \\ Input Array (left), Output Array (mid), Filter (right)}
\label{fig:convolution}
\end{figure}

Images are often represented by a three-dimensional array with dimensions height, width, and number of color channels. 
CNNs efficiently map image inputs to outputs by using weight-sharing filters, reducing the number of network parameters. 
A simple example of convolution is shown in~\Cref{fig:convolution}.
As the filter slides over the input array, a linear transform is applied to the group of array elements to produce the output. 
The stride of the filter, how far it is translated each step, dictates the size of the output array. 
The example in~\Cref{fig:convolution} uses a stride of 1.

 
CNNs achieved state-of-the-art performance on many visual processing tasks due to their ability to efficiently learn filters over varying spatial scales~\cite{alom2018}. 
One class of tasks is image-to-image translation, in which an input image is transformed such that some of its features match those of a target class of images (e.g. photograph to painting style transfer). 
We cast the problem of depth map generation as an image-to-image translation task, with the visual camera images as the input and depth maps as the output. 
In this context, the intensity of each pixel value of the depth map is indicative of the relative distance of the object in the scene.

A challenge to this approach is that monocular images contain no explicit information about the distance of objects in scene. 
While some relative distance information may be inferred from relational cues (e.g. object obfuscation), actual distance remains ambiguous in the 2D representation. 
Our approach seeks to resolve that ambiguity by allowing comparisons between sequential frames. 
This type of approach is known as depth-from-motion~\cite{sperling1994}. 
To allow this comparison in a deep neural network, we introduce recurrent convolutional cells to the network. 

Recurrent cells are neural network activation functions that maintain a latent state during the course of network execution. 
The input and the latent state are used to generate the recurrent cell output and update the latent state. 
We use the convolutional Gated Recurrent Unit (GRU) as our recurrent cell~\cite{Cho2014}.  
The convolutional GRU performs non-linear convolutional operations on inputs to generate outputs and update its latent state.
Prior work implementing a similar architecture has shown that the convolutional GRU cell performance meets or exceeds the performance of the similar convolutional Long Short-Term Memory (LSTM) cell while being simpler to implement and train~\cite{Toderici2017}. 

\begin{figure}[h]
\begin{center}
\includegraphics[width=0.9\textwidth]{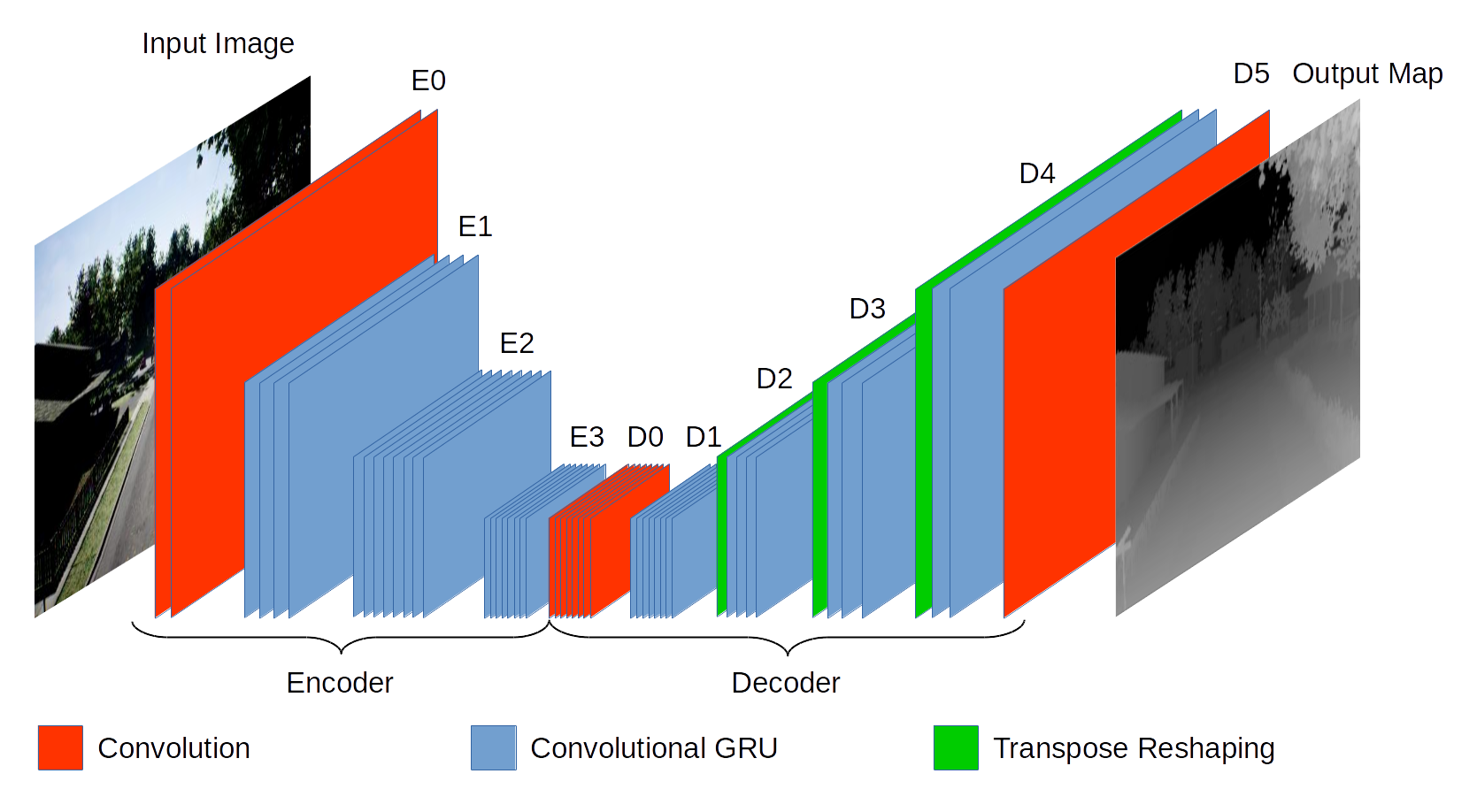}
\end{center}
\caption{Proposed neural network architecture}\label{fig:NN}
\end{figure}

Using GRU cells, we propose an autoencoder architecture as shown in~\Cref{fig:NN}. 
An autoencoder is composed of an encoder, bottleneck layer, and decoder. 
The encoder reduces the size of the layer outputs through recurrent strided convolutions until a minimum sized bottleneck layer (E3 in~\Cref{fig:NN}), and the decoder then increases the output size by reshaping some of the array elements in the depth dimension to the spatial dimensions. 
Unlike transpose convolution or interpolation, this reshaping method does not create artifacts in the output images because the network can directly learn the proper mapping across the reshaping.


\begin{table}[h]
\caption{Neural network design parameters}\label{table:NN}
\centering
\begin{tabular}{lrrrr} 
\toprule
\textbf{Layer} & \textbf{Filter Size} & \textbf{Stride} & \textbf{Depth} & \textbf{Activation} \\
\midrule
E0 & (3,3) & 2 & 64 & LReLU \\
E1 & (3,3) & 2 & 256 & GRU \\
E2 & (3,3) & 2 & 512 & GRU \\
E3 & (3,3) & 2 & 512 & GRU \\
\midrule
D0 & (1,1) & 1 & 512 & LReLU \\
\midrule
\multicolumn{5}{c}{Reshape} \\
\midrule
D1 & (3,3) & 1 & 512 & GRU \\
\midrule
\multicolumn{5}{c}{Reshape} \\
\midrule
D2 & (3,3) & 1 & 256 & GRU \\
\midrule
\multicolumn{5}{c}{Reshape} \\
\midrule
D3 & (3,3) & 1 & 256 & GRU \\
\midrule
\multicolumn{5}{c}{Reshape} \\
\midrule
D4 & (3,3) & 1 & 128 & GRU \\
D5 & (1,1) & 1 & 3 & tanh \\
\bottomrule
\end{tabular}
\end{table}

Our network architecture is described in~\Cref{table:NN}, where the $Layer$ corresponds to the matching labels in~\cref{fig:NN}, $Depth$ defines the number of filters applied at each layer, and $Activation$ defines the non-linear function used at each layer. 
GRU cells were used for all the recurrent activation functions. 
For layers E0 and D0, the leaky Rectified Linear Unit (LReLU) was used, which is defined as 
\begin{equation}
f(x) = \alpha \ \text{max}(0, x) + (1 - \alpha) \text{max}(0, -x)
\label{eq:LReLU}
\end{equation}
where $\alpha$ is the leak parameter which was defined as $\alpha = 0.1$ for all layers. 
Before each decoder layer, a reshaping upscales the spatial dimensions by a factor of two in both height and width.

The network was trained using supervised learning, which required matched pairs of camera images and true depth maps.
The loss function was the L1 norm of the difference between the generated depth map and the true depth map, which is defined as 
\begin{equation}
j(y, \hat{y}) = \| y - \hat{y}\|_1
\label{eq:L1}
\end{equation}
where $y$ is the true depth map and $\hat{y}$ is the depth map output by the network.
The network parameters were optimized using gradient-based optimization using the Adam optimizer~\cite{Kingma2014}. 
Our network was built and trained using the Tensorflow framework~(tensorflow.org).

At each optimization step, data was provided in mini-batches comprised of sub-sets of the complete training set. 
The full data set was composed of videos showing episodes of different trajectories in the simulated environment.
The complete episode videos were not used directly for training because recurrent neural networks are sensitive to the vanishing gradient problem~\cite{Pascanu2012}, and full-episodes were typically longer sequences than would be viable. 
Training sequences were instead generated as 32-frame sub-sequences from the full episodes.

During training, these mini-batches were constructed by uniformly sampling a starting frame from the complete set and constructing the sequence from the following frames. 
The loss used for the optimization step was the mean loss of the mini-batch, which is defined as 
\begin{equation}
J(Y, \hat{Y}) = \frac{1}{m} \sum_{i=1}^m \| y_i - \hat{y_i}\|_1
\label{eq:loss}
\end{equation}
where $Y$ is the set of true depth maps, $\hat{Y}$ is the set of depth maps generated by the network, and $m$ is the number of images in the mini-batch.
The training mini-batches are sequential, so the network can learn the depth-from-motion relationships in frame sequences.
Sampling the starting index stochastically decorrelates the optimization steps, stabilizing the training process. 

During operation of the network, the GRU latent states are continually updated each time a frame is passed through the graph and the output of the network for each frame is dependent upon its initial latent state.  
We draw our mini-batch samples stochastically, so we do not know the initial latent state corresponding to the given training sequence.
Initializing this latent state to zero during training can bias the training process. 
To overcome this, we introduce a burn-in period for each training update, where we construct an initialization sequence of the 32 images \emph{before} the selected start frame. 
We feed these through the network without including the error in the loss function, allowing the network to accumulate a latent state. 
The optimization step then proceeds with the training batch using the initialized network. 

\section{Experimental Setup}
\begin{figure}[h]
\begin{center}
\includegraphics[width=0.4\textwidth]{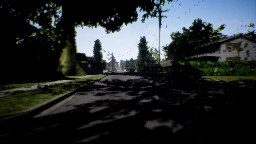}
\includegraphics[width=0.4\textwidth]{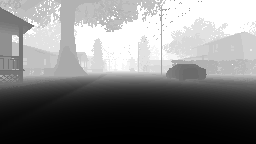}
\end{center}
\caption{(Left) Synthetic camera image; (Right) Corresponding depth map}\label{fig:AirSim}
\end{figure}

We ran a series of experiments to qualitatively and quantitatively evaluate the performance of our method using Microsoft's AirSim UAV simulator. 
We used AirSim to gather our training dataset as well as a separate dataset used for testing. 
The datasets consisted of images from an on-board synthetic camera along with its corresponding depth map. 
\Cref{fig:AirSim} shows an example image pair. 
Our training dataset has 1873 total image pairs in sequence, for which we use 80\% for training and hold out 20\% as a validation set for hyperparameter tuning.
These were gathered by manually piloting the AirSim car model through various trajectories in AirSim.
Our test dataset has 750 image pairs gathered from the same environment. 

In addition to evaluating our method, we implemented three baseline CNNs for comparison: Pix2Pix, CycleGAN, and multi-scale deep network. 
The first two baselines are commonly used for image-to-image translation problems. 
They implement Generative Adversarial Networks (GANs) to create outputs that are perceptually similar to the target image class. 
Pix2Pix is a conditional GAN developed for general image-to-image translation tasks~\cite{Isola2016}. 
An extension to Pix2Pix is CycleGAN, which not only learns to map an input image to an output image but ensures that the output image can be used to recreate the input image~\cite{Zhu2017}. 
The final method uses a multi-scale CNN, which first trains a coarse depth map generator and then trains a second network to refine the coarse depth map~\cite{Eigen2014}.
These networks were chosen because they have achieved state-of-the-art performance in various image translation tasks. 

We trained our Convolutional GRU network using stochastic sequential mini-batches for 10,000 epochs with an initial learning rate of $10^{-3}$. 
We trained and evaluated the baseline networks with the same data sets and hyperparameters determined by cross-validation for each network.
After training, we generated depth maps for the data in the test set and measured the pixel average mean-square error (MSE), pixel average absolute error (AE), and the average pixel Root Mean-Square Logistic Error (RMSLE) for all four models. 
All of the errors are an average over the individual pixel color channel values of the real and network generated depth images, labeled $d_{\text{real}}$ and $d_{\text{network}}$ respectively. 
Because our depth maps are in gray-scale, pixel values in all three color channels are identical.
Let $c$ be the total number of color channels each pixel has, let $n$ be the total number of pixels in the test set, and let $d^{(i,j)}_{\text{real}}$ and $d^{(i,j)}_{\text{network}}$ be the $i^{th}$ color channel of the $j^{th}$ pixel of the real and network generated depth maps with $i \in {1,...,c}$ and $j \in {1,...,n}$.
These error terms are defined as 
\begin{align}
\text{MSE} & = \frac{1}{c} \frac{1}{n} \sum_{i=1}^{c}\sum_{j=1}^{n}\big(d^{(i,j)}_{\text{real}} - d^{(i,j)}_{\text{network}}\big)^2 \\
\text{AE} & = \frac{1}{c} \frac{1}{n} \sum_{i=1}^{c}\sum_{j=1}^{n}\big\vert d^{(i,j)}_{\text{real}} - d^{(i,j)}_{\text{network}} \big\vert \\
\text{RMSLE} & = \sqrt{\frac{1}{c}\frac{1}{n} \sum_{i=1}^{c}\sum_{j=1}^{n}\big(\log(256-d^{(i)}_{\text{real}})-\log(256-d^{(i)}_{\text{network}})\big)^2}
\end{align}

While all of these metrics provide a measure of image accuracy, each of tends to weight accuracy of a different perceptual image quality. 
The MSE tends to emphasize the low-frequency content of an image caused by large objects.
AE tends to more heavily weight the high-frequency content of an image, such as object textures. 
An image that performs well in MSE but not in AE will often provide a blurred rendering of the true image. 
RMSLE captures accuracy of features relative to the intensity of the feature pixels. 
For example, an error of 10 on of a feature whose average pixel value is 200 will be penalized far less than an error of 10 on a feature whose average pixel value is 20. 
In this way, RMSLE can be interpreted as a perceptually weighted loss.  

\begin{figure}[h]
\begin{center}
\includegraphics[width=0.5 \textwidth]{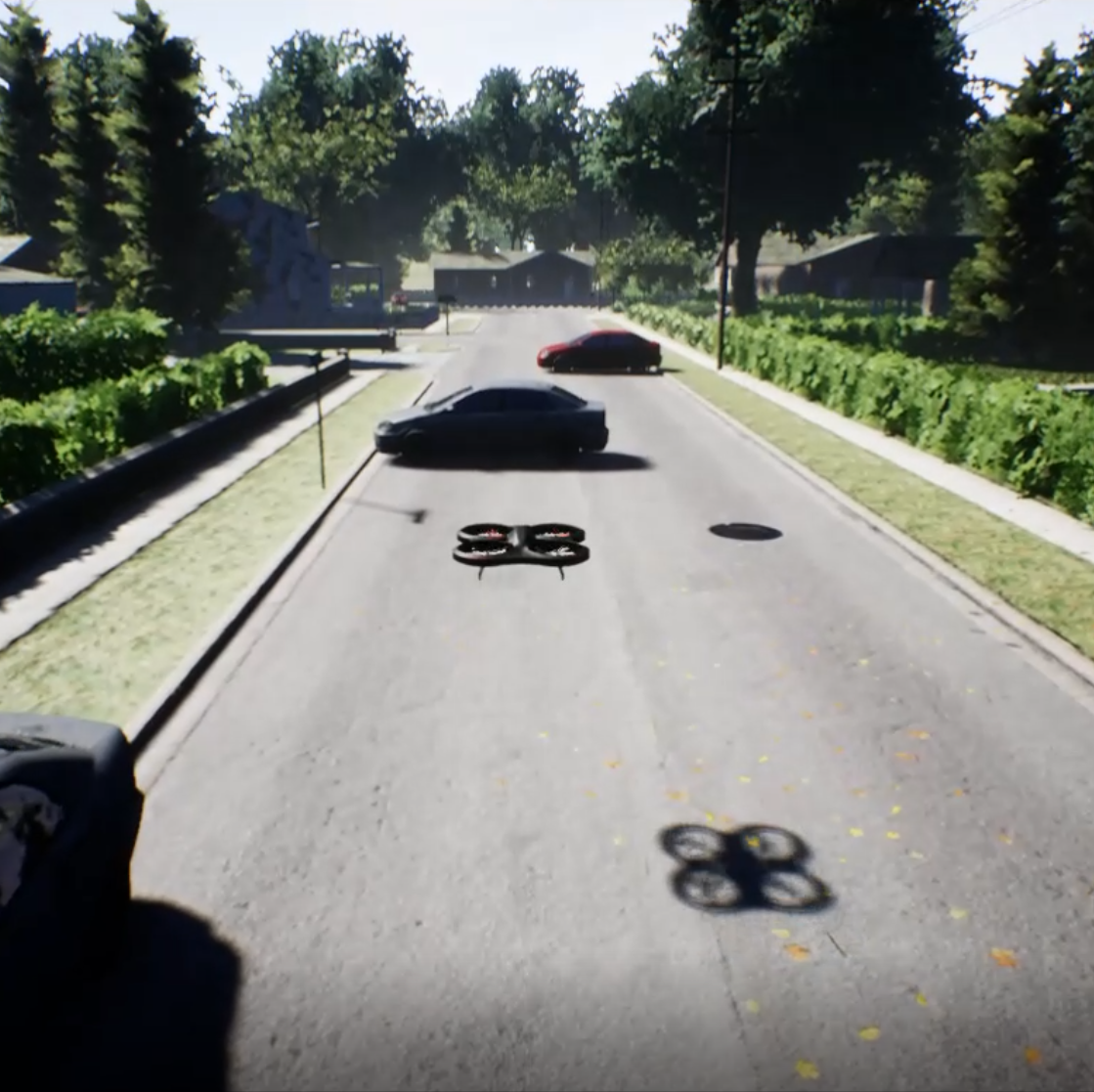}
\hspace{1cm}
\begin{tikzpicture}

	\node[anchor=south west,inner sep=0] at (1.5,2.7) {\begin{rotate}{90}\scalebox{0.15}		
    		{\begin{tikzpicture}
\definecolor{mycolor}{RGB}{72,130,206}
\draw [very thick,rounded corners=20, fill=mycolor] (0,0) -- (3,0) --(3,5.5) -- (0,5.5) -- cycle ;

\draw (2.35,5.2) -- (2.65,5.);
\draw  [fill=white] (2.35,5.2) to[bend left=100] (2.65,5.);

\draw (0.65,5.2) -- (0.35,5.);
\draw  [fill=white] (0.65,5.2) to[bend right=100] (0.35,5.);

\draw [fill=white] (0.75,3.5) -- (2.25,3.5) -- (2.65,4.) to[out=155,in=25] (0.35,4.) -- cycle;

\draw [fill=white] (0.75,1.5) -- (2.25,1.5) -- (2.65,1) to[out=-155,in=-25] (0.35,1) --cycle; 

\draw (2.85,1.2) -- (2.85,3.7);
\draw  [fill=white] (2.85,1.2) to[bend left=30] (2.85,3.7);

\draw [fill=white] (0.15,1.2) -- (0.15,3.7);
\draw  [fill=white] (0.15,1.2) to[bend right=30] (0.15,3.7);

\draw [thick, rounded corners=12,fill=white] (3.,4.5) -- (3.5,4.) -- (3,4);

\draw [thick, rounded corners=12, fill=white] (0.,4.5) -- (-0.5,4.) -- (0,4);

\end{tikzpicture}}\end{rotate}};
    \node[anchor=south west,inner sep=0] at (2.05,4.55) {\begin{rotate}{90}\scalebox{0.15}
    		{\begin{tikzpicture}
\definecolor{mycolor}{RGB}{165,40,40}
\draw [very thick,rounded corners=20,fill=mycolor] (0,0) -- (3,0) --(3,5.5) -- (0,5.5) -- cycle ;

\draw (2.35,5.2) -- (2.65,5.);
\draw  [fill=white] (2.35,5.2) to[bend left=100] (2.65,5.);

\draw (0.65,5.2) -- (0.35,5.);
\draw  [fill=white] (0.65,5.2) to[bend right=100] (0.35,5.);

\draw [fill=white] (0.75,3.5) -- (2.25,3.5) -- (2.65,4.) to[out=155,in=25] (0.35,4.) -- cycle;

\draw [fill=white] (0.75,1.5) -- (2.25,1.5) -- (2.65,1) to[out=-155,in=-25] (0.35,1) --cycle; 

\draw (2.85,1.2) -- (2.85,3.7);
\draw  [fill=white] (2.85,1.2) to[bend left=30] (2.85,3.7);

\draw [fill=white] (0.15,1.2) -- (0.15,3.7);
\draw  [fill=white] (0.15,1.2) to[bend right=30] (0.15,3.7);

\draw [thick, rounded corners=12,fill=white] (3.,4.5) -- (3.5,4.) -- (3,4);

\draw [thick, rounded corners=12, fill=white] (0.,4.5) -- (-0.5,4.) -- (0,4);

\end{tikzpicture}}\end{rotate}};
            
\begin{axis}[
	xlabel=Crossrange (m), 
	ylabel=Downrange (m),
	height=8.7cm,
	width=4.0cm,
	xmin=-5.0, xmax=5,
	ymin=-10.0, ymax=65.0]
\addplot +[black, solid, thick, mark=none] 
	table[x=x, y=y, col sep=comma]{figs/TrajsTrue/Traj_Header_00.csv};
\end{axis}

\node[text width=3cm] at (2.4,0.6) {Start};
\node[text width=3cm] at (2.3,6.7) {Finish};

\end{tikzpicture}
\end{center}
\caption{A snapshot of the Airsim simulation (left) and a top-view of the obstacle course with the expected trajectory (right)} \label{fig:AirsimCourse}
\end{figure}
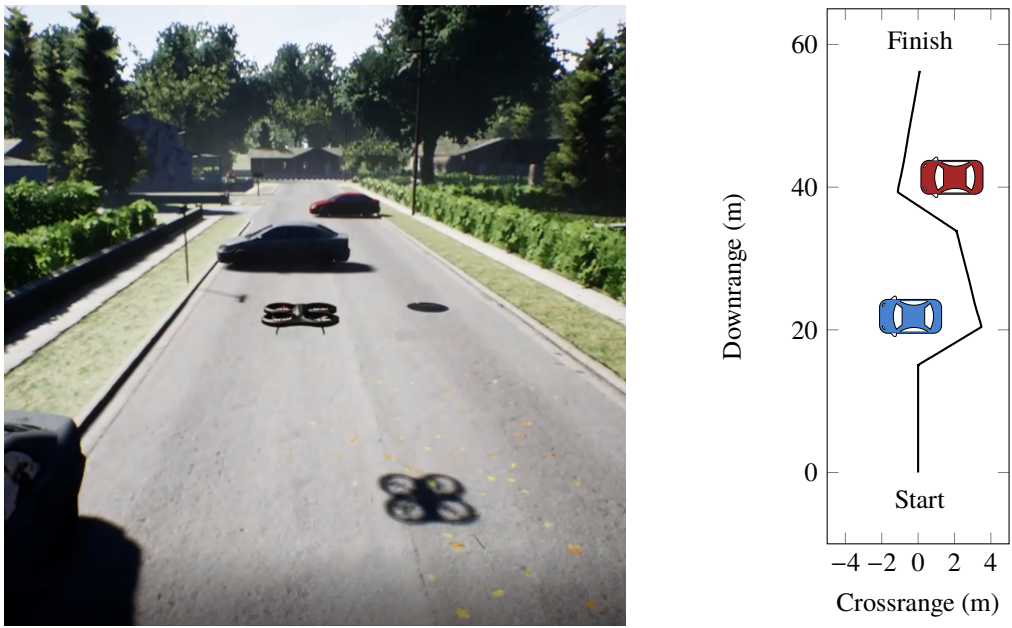

In addition to the test set comparison, we also evaluated the effectiveness of each method for obstacle avoidance.
To do so, we created an obstacle course in AirSim by placing cars in the road that block an unmanned aircraft from reaching its goal. 
To avoid the cars and reach the goal, the aircraft uses the depth prediction models to assess when it should stop before flying into a car as well as how much turning is needed to avoid the car. 
Cars parked perpendicularly in the road were not present in the environment from which the training data was gathered. 
In addition, no collisions with cars were seen in the training set, so this experiment evaluates the ability of the networks to generalize to new scenarios with new obstacles.  
\Cref{fig:AirsimCourse} shows an example trajectory as well as a top-down view of the trajectory flown when using the true depth map to guide the aircraft.

\section{Results}

\begin{figure}[h]
\begin{center}
\includegraphics[width=0.9 \textwidth]{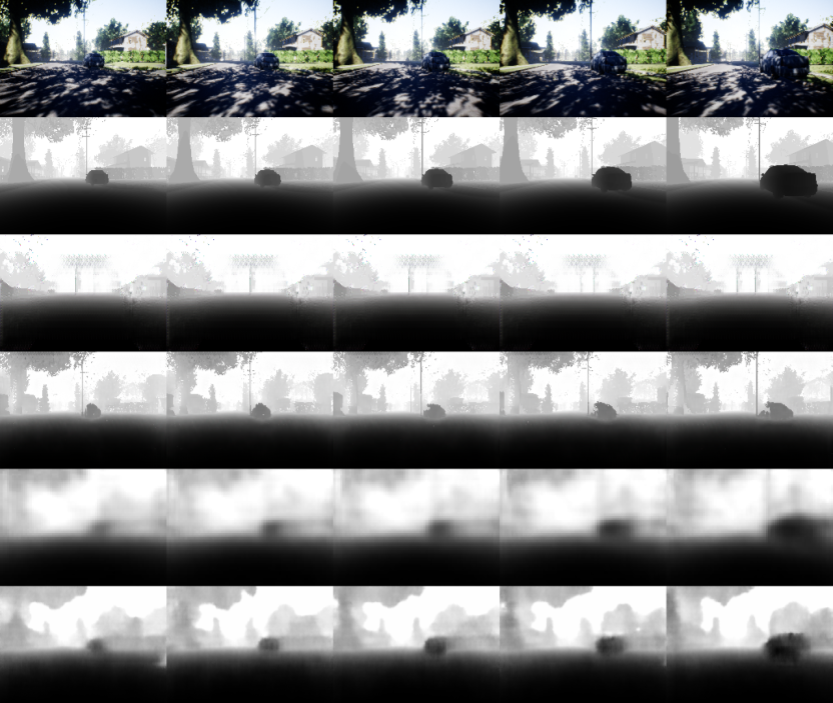}
\end{center}
\caption{Example frame sequences. Rows (top to bottom): Visual Input Image, True Depth Map, Pix2Pix, Cycle GANs, Multi-scale CNN, Convolutional GRU}\label{fig:results}
\end{figure}

We qualitatively assessed the depth maps generated from each network. 
\Cref{fig:results} shows a sample sequence from the different networks. 
As can be seen, Pix2Pix generated a depth map for each input frame which tended not to vary with inputs. 
It is likely that this depth map represents the average scene observed in the dataset. 
At first glance, Cycle GAN appears to provide the clearest resolution. 
Upon further inspection, it can be seen that, while visually detailed, the actual depth values are typically inaccurate. 
This is likely an effect of the cycle GAN loss which prioritizes the visual perceptual qualities of the image (presence of textures, shapes, etc.) over the agreement with the target image. 
This effect can be clearly seen in the frame-to-frame evolution of the GAN depth maps.
The car is initially textured as a bush and continues to morph in shape throughout the progression. 
Additionally, several background features appear and disappear throughout the sequence, and notably the house is absent in all frames. 
The multi-scale CNN captures the very low-frequency features of the image (e.g. the car), though the blur renders most of the background indistinguishable. 
Our proposed network appears to provide the best resolution of details while accurately matching the values of the true depth map features.  

\Cref{table:results} provides a quantitative comparison between our method and the baselines. 
As can be seen, our recurrent method outperforms the baseline methods on all metrics.
In agreement with the qualitative assessment, Pix2Pix has the worst quantitative performance in all categories. 
Our qualitative assessment of cycle GAN is supported by the qualitative results. 
While features in the depth map were clearly resolved, the accuracy of the resulting maps is fairly poor, performing worse than the multi-scale CNN and the Convolutional GRU. 
The multi-scale CNN performance nearly matches the performance of the convolutional GRU in MSE, with only a $4.5\%$ difference. 
This fits with our qualitative assessment, as the multi-scale CNN maps were observed to accurately capture blurry representations of large features in the scene. 
The absolute error, however, is $20.0\%$ higher than the convolutional GRU.
This is again consistent with our observations, as the convolutional GRU features were much more clearly resolved.  
The improved performance of the convolutional GRU over the baseline methods is likely due to the ability of the convolutional GRU to make implicit depth-from-motion estimates from the image sequences, while the baseline methods are only consider a single frame per estimate. 

\begin{table}[h!]
\caption{Test set results}\label{table:results}
\centering
\begin{tabular}{lrrr} 
\toprule
{Network} & {MSE} & {AE} & {RMSLE} \\
\midrule
Pix2Pix & 3534.6 & 36.41 & 0.419 \\
CycleGAN & 720.8 & 14.63 & 0.242 \\
Multi-scale CNN & 478.6 & 12.83 & 0.234 \\
\textbf{Convolutional GRU} & \textbf{457.4} & \textbf{10.72} & \textbf{0.081} \\
\bottomrule
\end{tabular}
\end{table}

\begin{figure}[h]
\begin{center}
\begin{tikzpicture}

	\node[anchor=south west,inner sep=0] at (1.5,2.7) {\begin{rotate}{90}\scalebox{0.15}		
    		{\begin{tikzpicture}
\definecolor{mycolor}{RGB}{72,130,206}
\draw [very thick,rounded corners=20, fill=mycolor] (0,0) -- (3,0) --(3,5.5) -- (0,5.5) -- cycle ;

\draw (2.35,5.2) -- (2.65,5.);
\draw  [fill=white] (2.35,5.2) to[bend left=100] (2.65,5.);

\draw (0.65,5.2) -- (0.35,5.);
\draw  [fill=white] (0.65,5.2) to[bend right=100] (0.35,5.);

\draw [fill=white] (0.75,3.5) -- (2.25,3.5) -- (2.65,4.) to[out=155,in=25] (0.35,4.) -- cycle;

\draw [fill=white] (0.75,1.5) -- (2.25,1.5) -- (2.65,1) to[out=-155,in=-25] (0.35,1) --cycle; 

\draw (2.85,1.2) -- (2.85,3.7);
\draw  [fill=white] (2.85,1.2) to[bend left=30] (2.85,3.7);

\draw [fill=white] (0.15,1.2) -- (0.15,3.7);
\draw  [fill=white] (0.15,1.2) to[bend right=30] (0.15,3.7);

\draw [thick, rounded corners=12,fill=white] (3.,4.5) -- (3.5,4.) -- (3,4);

\draw [thick, rounded corners=12, fill=white] (0.,4.5) -- (-0.5,4.) -- (0,4);

\end{tikzpicture}}\end{rotate}};
    \node[anchor=south west,inner sep=0] at (2.05,4.55) {\begin{rotate}{90}\scalebox{0.15}
    		{\begin{tikzpicture}
\definecolor{mycolor}{RGB}{165,40,40}
\draw [very thick,rounded corners=20,fill=mycolor] (0,0) -- (3,0) --(3,5.5) -- (0,5.5) -- cycle ;

\draw (2.35,5.2) -- (2.65,5.);
\draw  [fill=white] (2.35,5.2) to[bend left=100] (2.65,5.);

\draw (0.65,5.2) -- (0.35,5.);
\draw  [fill=white] (0.65,5.2) to[bend right=100] (0.35,5.);

\draw [fill=white] (0.75,3.5) -- (2.25,3.5) -- (2.65,4.) to[out=155,in=25] (0.35,4.) -- cycle;

\draw [fill=white] (0.75,1.5) -- (2.25,1.5) -- (2.65,1) to[out=-155,in=-25] (0.35,1) --cycle; 

\draw (2.85,1.2) -- (2.85,3.7);
\draw  [fill=white] (2.85,1.2) to[bend left=30] (2.85,3.7);

\draw [fill=white] (0.15,1.2) -- (0.15,3.7);
\draw  [fill=white] (0.15,1.2) to[bend right=30] (0.15,3.7);

\draw [thick, rounded corners=12,fill=white] (3.,4.5) -- (3.5,4.) -- (3,4);

\draw [thick, rounded corners=12, fill=white] (0.,4.5) -- (-0.5,4.) -- (0,4);

\end{tikzpicture}}\end{rotate}};
            
\begin{axis}[
	xlabel=Crossrange (m), 
	ylabel=Downrange (m),
	height=8.7cm,
	width=4.0cm,
	xmin=-5.0, xmax=5,
	ymin=-10.0, ymax=65.0,
    cycle list/Set2]
   
\foreach \i in {0,1,...,29}
{
	\addplot +[solid, mark=none] 
	table[x=x, y=y, col sep=comma]{figs/TrajsMern/Traj_Header_\i.csv};
}

\end{axis}

\node[text width=3cm] at (2.4,0.6) {Start};
\node[text width=3cm] at (2.3,6.7) {Finish};
\end{tikzpicture}
\hspace{-0.5cm}
\begin{tikzpicture}

	\node[anchor=south west,inner sep=0] at (1.5,2.7) {\begin{rotate}{90}\scalebox{0.15}		
    		{\begin{tikzpicture}
\definecolor{mycolor}{RGB}{72,130,206}
\draw [very thick,rounded corners=20, fill=mycolor] (0,0) -- (3,0) --(3,5.5) -- (0,5.5) -- cycle ;

\draw (2.35,5.2) -- (2.65,5.);
\draw  [fill=white] (2.35,5.2) to[bend left=100] (2.65,5.);

\draw (0.65,5.2) -- (0.35,5.);
\draw  [fill=white] (0.65,5.2) to[bend right=100] (0.35,5.);

\draw [fill=white] (0.75,3.5) -- (2.25,3.5) -- (2.65,4.) to[out=155,in=25] (0.35,4.) -- cycle;

\draw [fill=white] (0.75,1.5) -- (2.25,1.5) -- (2.65,1) to[out=-155,in=-25] (0.35,1) --cycle; 

\draw (2.85,1.2) -- (2.85,3.7);
\draw  [fill=white] (2.85,1.2) to[bend left=30] (2.85,3.7);

\draw [fill=white] (0.15,1.2) -- (0.15,3.7);
\draw  [fill=white] (0.15,1.2) to[bend right=30] (0.15,3.7);

\draw [thick, rounded corners=12,fill=white] (3.,4.5) -- (3.5,4.) -- (3,4);

\draw [thick, rounded corners=12, fill=white] (0.,4.5) -- (-0.5,4.) -- (0,4);

\end{tikzpicture}}\end{rotate}};
    \node[anchor=south west,inner sep=0] at (2.05,4.55) {\begin{rotate}{90}\scalebox{0.15}
    		{\begin{tikzpicture}
\definecolor{mycolor}{RGB}{165,40,40}
\draw [very thick,rounded corners=20,fill=mycolor] (0,0) -- (3,0) --(3,5.5) -- (0,5.5) -- cycle ;

\draw (2.35,5.2) -- (2.65,5.);
\draw  [fill=white] (2.35,5.2) to[bend left=100] (2.65,5.);

\draw (0.65,5.2) -- (0.35,5.);
\draw  [fill=white] (0.65,5.2) to[bend right=100] (0.35,5.);

\draw [fill=white] (0.75,3.5) -- (2.25,3.5) -- (2.65,4.) to[out=155,in=25] (0.35,4.) -- cycle;

\draw [fill=white] (0.75,1.5) -- (2.25,1.5) -- (2.65,1) to[out=-155,in=-25] (0.35,1) --cycle; 

\draw (2.85,1.2) -- (2.85,3.7);
\draw  [fill=white] (2.85,1.2) to[bend left=30] (2.85,3.7);

\draw [fill=white] (0.15,1.2) -- (0.15,3.7);
\draw  [fill=white] (0.15,1.2) to[bend right=30] (0.15,3.7);

\draw [thick, rounded corners=12,fill=white] (3.,4.5) -- (3.5,4.) -- (3,4);

\draw [thick, rounded corners=12, fill=white] (0.,4.5) -- (-0.5,4.) -- (0,4);

\end{tikzpicture}}\end{rotate}};
            
\begin{axis}[
	xlabel=Crossrange (m), 
	height=8.7cm,
	width=4.0cm,
	xmin=-5.0, xmax=5,
	ymin=-10.0, ymax=65.0,
    cycle list/Set2]
    
\foreach \i in {0,1,...,29}
{
	\addplot +[solid, mark=none] 
	table[x=x, y=y, col sep=comma]{figs/TrajsJulian/Traj_Header_\i.csv};
}


\end{axis}
\node[text width=3cm] at (2.4,0.6) {Start};
\node[text width=3cm] at (2.3,6.7) {Finish};
\end{tikzpicture}
\hspace{-0.5cm}
\begin{tikzpicture}

	\node[anchor=south west,inner sep=0] at (1.5,2.7) {\begin{rotate}{90}\scalebox{0.15}		
    		{\begin{tikzpicture}
\definecolor{mycolor}{RGB}{72,130,206}
\draw [very thick,rounded corners=20, fill=mycolor] (0,0) -- (3,0) --(3,5.5) -- (0,5.5) -- cycle ;

\draw (2.35,5.2) -- (2.65,5.);
\draw  [fill=white] (2.35,5.2) to[bend left=100] (2.65,5.);

\draw (0.65,5.2) -- (0.35,5.);
\draw  [fill=white] (0.65,5.2) to[bend right=100] (0.35,5.);

\draw [fill=white] (0.75,3.5) -- (2.25,3.5) -- (2.65,4.) to[out=155,in=25] (0.35,4.) -- cycle;

\draw [fill=white] (0.75,1.5) -- (2.25,1.5) -- (2.65,1) to[out=-155,in=-25] (0.35,1) --cycle; 

\draw (2.85,1.2) -- (2.85,3.7);
\draw  [fill=white] (2.85,1.2) to[bend left=30] (2.85,3.7);

\draw [fill=white] (0.15,1.2) -- (0.15,3.7);
\draw  [fill=white] (0.15,1.2) to[bend right=30] (0.15,3.7);

\draw [thick, rounded corners=12,fill=white] (3.,4.5) -- (3.5,4.) -- (3,4);

\draw [thick, rounded corners=12, fill=white] (0.,4.5) -- (-0.5,4.) -- (0,4);

\end{tikzpicture}}\end{rotate}};
    \node[anchor=south west,inner sep=0] at (2.05,4.55) {\begin{rotate}{90}\scalebox{0.15}
    		{\begin{tikzpicture}
\definecolor{mycolor}{RGB}{165,40,40}
\draw [very thick,rounded corners=20,fill=mycolor] (0,0) -- (3,0) --(3,5.5) -- (0,5.5) -- cycle ;

\draw (2.35,5.2) -- (2.65,5.);
\draw  [fill=white] (2.35,5.2) to[bend left=100] (2.65,5.);

\draw (0.65,5.2) -- (0.35,5.);
\draw  [fill=white] (0.65,5.2) to[bend right=100] (0.35,5.);

\draw [fill=white] (0.75,3.5) -- (2.25,3.5) -- (2.65,4.) to[out=155,in=25] (0.35,4.) -- cycle;

\draw [fill=white] (0.75,1.5) -- (2.25,1.5) -- (2.65,1) to[out=-155,in=-25] (0.35,1) --cycle; 

\draw (2.85,1.2) -- (2.85,3.7);
\draw  [fill=white] (2.85,1.2) to[bend left=30] (2.85,3.7);

\draw [fill=white] (0.15,1.2) -- (0.15,3.7);
\draw  [fill=white] (0.15,1.2) to[bend right=30] (0.15,3.7);

\draw [thick, rounded corners=12,fill=white] (3.,4.5) -- (3.5,4.) -- (3,4);

\draw [thick, rounded corners=12, fill=white] (0.,4.5) -- (-0.5,4.) -- (0,4);

\end{tikzpicture}}\end{rotate}};
            
\begin{axis}[
	xlabel=Crossrange (m), 
	height=8.7cm,
	width=4.0cm,
	xmin=-5.0, xmax=5,
	ymin=-10.0, ymax=65.0,
    cycle list/Set2]
  
\foreach \i in {0,1,...,29}
{
	\addplot +[solid, mark=none] 
	table[x=x, y=y, col sep=comma]{figs/TrajsTompa/Traj_Header_\i.csv};
}
    
\end{axis}

\node[text width=3cm] at (2.4,0.6) {Start};
\node[text width=3cm] at (2.3,6.7) {Finish};
\end{tikzpicture}
\end{center}
\caption{Thirty simulated trajectories using a Convolutional GRU (left), Multi-scale CNN (middle), and Cycle GAN (right)}\label{fig:Trajs}
\end{figure}
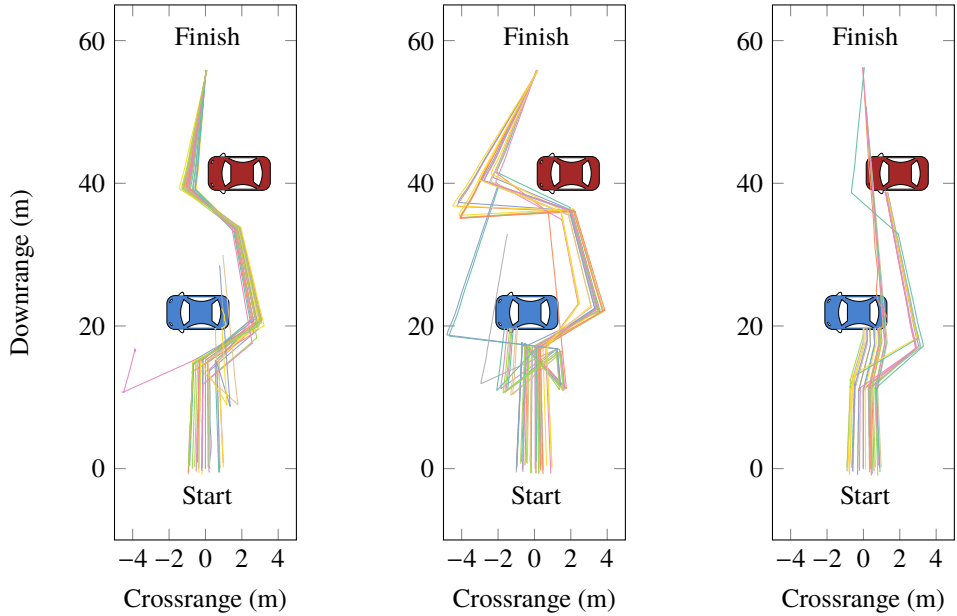

The Pix2Pix depth model was not evaluated with the obstacle course because it preformed poorly with the test image set. 
For the other three models, 30 simulations were conducted with randomized starting locations. 
\Cref{fig:Trajs} shows a top-down view of the 30 simulated trajectories flown with each method. 
In these cases where the trajectories intersect the bottom car, the aircraft flies over the trunk of the car. 
The simulation results are summarized in \cref{table:sim}.

The Convolutional GRU performs the best and stops before the cars at a consistent distance. 
A few times the aircraft turns farther than needed because the predicted depth map does not show an open path forward until facing almost the opposite direction. 
The aircraft is able to reach the goal in 25 of the 30 simulations and only crashes into a car twice. 
The remaining three trajectories timed out because the aircraft took wrong turns and could not reach the goal in the given time.
In some trials, the aircraft pitched significantly to stop, and the depth estimate noise caused the car to be lost. 
This is likely due to these types of maneuvers not being seen in the training data.
After the aircraft remained stationary for a few moments, the car was re-acquired and the travel resumed. 

The multi-scale CNN performs the second best in the simulations. 
The aircraft reaches the finish in 19 of the 30 simulations and crashes into a car 8 times. 
The mutli-scale CNN depth predictions are not as well resolved as with the Convolutional GRU, and as a result the depth predictions are not as accurate or reliable. 
Finally, the Cycle GAN model performs poorly. The aircraft reaches the finish in only 6 of the 30 simulations and crashes into a car in the remaining 24 trajectories. 
Although Cycle GAN predictions often seem the most visually detailed, they are often inaccurate. 
The resulting depth predictions do not get low enough to alert the vehicle that a car is in the path of the aircraft. 
As the aircraft approaches the car, the car seems to blend in with the road, which makes avoiding the car difficult.

\begin{table}[h!]
\caption{Results from 30 simulations}\label{table:sim}
\centering
\begin{tabular}{lrr} 
\toprule
{Network} & {Finishes} & {Crashes} \\
\midrule
CycleGAN & 6 & 24 \\
Multi-scale CNN & 19 & 8 \\
\textbf{Convolutional GRU} & \textbf{25} & \textbf{2} \\
\bottomrule
\end{tabular}
\end{table}

\section{Conclusion}
In this work, we presented a novel method to estimate object distances from a simple monocular camera using recurrent convolutional neural networks. 
We proposed a neural network architecture and design based on convolutional Gated Recurrent Units and a method to train the network using sequential stochastic mini-batch training. 
We also introduced hidden-state burn-in to reduce the bias induced by the stochastic training process. 

We demonstrated the effectiveness of this approach in a simulated environment. 
Our approach quantitatively outperformed state-of-the-art convolutional neural style transfer methods in three common objective quality metrics. 
We showed that with only an on-board monocular camera, the aircraft was able to resolve object depth with sufficient accuracy to avoid obstacles. 
Future work will investigate the effectiveness of this method in different visual environments and with moving obstacles.

While our recurrent network was able to outperform previous methods, the network architecture was relatively simple. 
Future work will explore the inclusion of architectural features of more complex style-transfer networks while retaining the recurrent units to improve the resolution of depth features. 
Currently, training is conducted with a simple L1 loss function. 
Loss functions tailored to specific depth map use cases, such as sense-and-avoid or SLAM, will also be investigated. 
Additionally, incorporating vehicle telemetry (velocity, orientation, etc.) as an additional network input will be explored to improve depth-from-motion estimation. 

\bibliography{bib}

\end{document}